\documentclass[10pt,twocolumn,letterpaper]{article}

\usepackage{cvpr}
\usepackage{times}
\usepackage{epsfig}
\usepackage{graphicx}
\usepackage{amsmath}
\usepackage{amssymb}
\usepackage{caption}
\usepackage{float}
\usepackage{bm}

\newcommand{\G}{G\xspace}
\newcommand{\D}{D\xspace}
\newcommand{\E}{E\xspace}

\usepackage[pagebackref=true,breaklinks=true,letterpaper=true,colorlinks,bookmarks=false]{hyperref}

\cvprfinalcopy 

\def\cvprPaperID{1038} 

\begin{document}

\title{Latent Filter Scaling for Multimodal Unsupervised Image-to-Image Translation}

\author{Yazeed Alharbi\\
King Abdullah University for Science and Technology (KAUST)\\
{\tt\small yazeed.alharbi@kaust.edu.sa}
\and
Neil Smith\\
KAUST\\
\and
Peter Wonka\\
KAUST\\
}

\twocolumn[{%
\renewcommand\twocolumn[1][]{#1}%
\maketitle
\begin{center}
    \centering
 \includegraphics[width=1\linewidth]{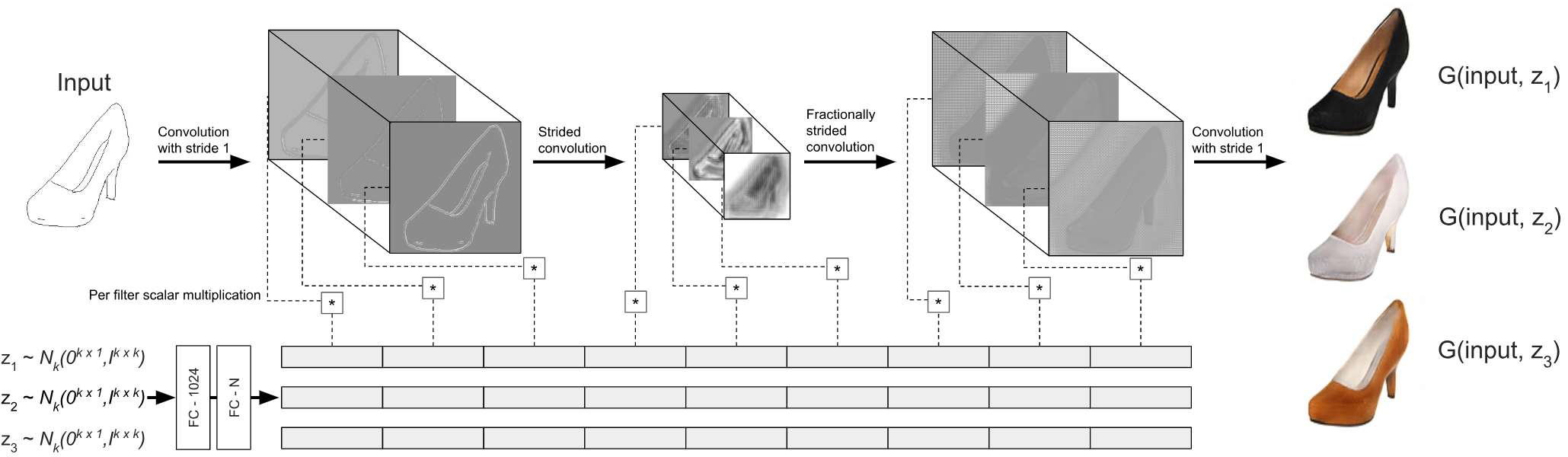}
\captionof{figure}{An illustration of latent filter scaling. Input noise codes are mapped to the filters of the network, instead of injection into the input image.}
\end{center}%
}]


\begin{abstract}
In multimodal unsupervised image-to-image translation tasks, the goal is to translate an image from the source domain to many images in the target domain. We present a simple method that produces higher quality images than current state-of-the-art while maintaining the same amount of multimodal diversity. Previous methods follow the unconditional approach of trying to map the latent code directly to a full-size image. This leads to complicated network architectures with several introduced hyperparameters to tune. By treating the latent code as a modifier of the convolutional filters, we produce multimodal output while maintaining the traditional Generative Adversarial Network (GAN) loss and without additional hyperparameters. The only tuning required by our method controls the tradeoff between variability and quality of generated images. Furthermore, we achieve disentanglement between source domain content and target domain style for free as a by-product of our formulation. We perform qualitative and quantitative experiments showing the advantages of our method compared with the state-of-the art on multiple benchmark image-to-image translation datasets. 
\end{abstract}

\section{Introduction}
Recently, GANs have emerged as a promising research field \cite{ganPaper, wassGAN, ganPrinciples, stackedGAN, unrolledGAN}. The use of adversarial training has shown to be effective in many computer vision tasks. Generative networks can be trained with reasonable success to produce realistic images of humans and objects \cite{infoGAN, DCGAN, pyramidGAN}. \par
In the unconditional generation task, the aim is to map a randomly generated low-dimensional code to a realistic image. After training, the entries of the input code should control sources of variation in the output image. There is an effort in the community to guarantee certain desirable qualities about this mapping. For example, InfoGAN \cite{infoGAN} aims to enhance the interpretability of changing the latent code through disentanglement. Other methods aim for a mapping where the interpolation between two latent codes results in an image that is semantically interpolated between the images resulting from the two codes. Additionally, some methods aim for an application-specific disentangled mapping, where changing certain parts of the input code will correspond to changes in certain parts of the output images \cite{ganerated, personimage}. In the case of pedestrian image generation, the disentanglement might allow the user to change the pedestrian appearance independently of the background. \par
 In the literature, image-to-image translation is handled similarly to unconditional generation, where the latent vector is directly mapped to an image. In order to generate many images in the target domain for each input image, previous methods \cite{BicycleGAN,MUNIT,Diverse} make use of an input latent code in addition to autoencoding, adding multiple terms to the loss. The intuitive idea behind our contribution is that in image-to-image translation tasks we often start with an image and the aim is to use the latent vector to control local changes in the input image. Therefore, we opt for treating the latent vector as a modifier of the network's filters, instead of taking the unconditional approach of treating the latent vector as encoded data.\par
\textbf{Contributions:} Our contribution is a method for multimodal unsupervised image-to-image translation that is simple, competitive with the state-of-the-art, and provides disentanglement of the latent codes from the input images. This is achieved without modifying standard GAN loss and with minimal overhead. Our method achieves state-of-the-art quality and diversity on multiple image-to-image translation datasets, while maintaining the simplicity and generality of standard GAN loss and architecture. To the best of our knowledge, our method is the first method that does not require autoencoding or reconstruction losses for latent codes or images. Our method essentially prevents mode collapse in image-to-image GANs while enabling a larger degree of freedom in the quality-variability tradeoff. In contrast, previous methods include several losses in addition to the GAN loss, each with a new hyperparameter to tune.\par
\textbf{Notation:} In this paper, GANs that take an image from one domain and produce an image in another domain will be referred to as image-to-image translation GANs. If paired data are used, the GAN will be referred to as supervised. It will be referred to as unsupervised if the images from the two domains are not paired. Finally, image-to-image translation GANs that produce a single image will be referred to as deterministic or unimodal, while multimodal ones make use of an input latent vector in addition to the input image to produce many outputs. \par

\section{Related Work}
For paired data, conditional adversarial networks \cite{pairedCondGAN, highresCGAN} show reasonable results that are applicable to almost any dataset. However, requiring paired data is a major limitation. \par
Many works in literature search for ways to produce similar results without using pair information \cite{MUNIT, Diverse, AugmentedCYCLEGAN}. Zhu \etal \cite{CycleGAN} demonstrate  that unimodal unpaired image-to-image translation tasks are under-constrained, since there might be many possible translations in the output space. CycleGAN \cite{CycleGAN} is one of the most successful approaches to handling this issue. Instead of training one network that maps source domain images to target domain images, they propose adding another network to map domain images to source images forming a cycle. They argue that imposing this cycle-consistency loss produces better image quality. \par
While there are acceptable results for unconditional generation GANs and unimodal image-to-image translation GANs, the multimodal case is an open area of research. The success of GANs in generating diverse output images from scratch does not naturally extend to image-to-image translation problems. Diversity in the conditional case, where an image is given and the goal is to translate it into another class of images, is proving to be more challenging. Unlike the unconditional case, where the latent vector can be simply mapped to a full size image, the conditional case requires using both the latent vector and the input image. Simple concatenation of the latent vector with the image often leads to deterministic output, where the network learns to ignore the latent code. This is the problem that most conditional image-to-image translation papers tackle: preserving original image structure, and preserving influence and variability of the latent vector. In other words, given an input image and a latent vector, the aim is to change the appearance using the latent vector, such that different latent vectors produce different target domain images. \par
Earlier proposals for multimodal output are largely limited in capacity and application\cite{MADGAN, pixelNN, cascadeGAN}. In the case of PixelNN \cite{pixelNN}, multi-modality is achieved by providing some form of the output image to the GAN. For example, a low resolution image or a normal map must be fed to the GAN, and given different low resolution images different outputs will be produced. On the other hand, MADGAN \cite{MADGAN} employs multiple generators with the obvious limitation of having a constant and discrete number of generated target domain modes. Cascaded Refinement Networks \cite{cascadeGAN} have the same limitation of generating a constant discrete number of outputs at test time.\par
Previous methods \cite{BicycleGAN, MUNIT, Diverse, AugmentedCYCLEGAN} follow the unconditional approach for achieving conditional multi-modality. In the unconditional case, the latent code is often interpreted as a compressed version of the output image. After training an unconditional image-to-image translation GAN, it is often found that entries in the input noise vector correspond to semantic labels of the output images. For example, on MNIST, changing one entry in the input vector might change thickness or slope of a digit. In a sense, unconditional image generation is an inverse problem where the goal is to enforce the correspondence between latent code entries and semantic variations in the resulting image. Following this approach for image-to-image GANs, previous methods are often forced to compress the input image until concatenation with the low-dimensional code is meaningful. One of the more successful ways to achieve this is the disentanglement approach where the compression is regularized such that each image can be described by two components: the latent code which is domain specific, and a content code which is shared between source and target domain \cite{MUNIT, Diverse}. \par
One of the earlier methods for multimodal image-to-image translation is BicycleGAN \cite{BicycleGAN}. BicycleGAN combines two existing loss cycles: one to encourage latent code diversity and one to encourage faithfulness to ground truth images. One limitation of BicycleGAN is requiring paired images. This raises a new problem of producing multimodal output in an unsupervised fashion without requiring pair information. Recent methods \cite{MUNIT, Diverse} handle the problem by disentangling the input image into content and style. Content is assumed to be shared between the input and output distributions, while style is domain-specific. Then, to translate an image, the content code is computed and is concatenated with a style code sampled from target distribution. While augmented CycleGAN \cite{AugmentedCYCLEGAN} does not disentangle input images, they still follow the common approach of injecting the latent code somewhere in the network, and adding loss terms to ensure the latent code is meaningful and diverse. \par

While disentanglement can be desirable in some graphics problems where there is need to manipulate certain parts of the image independently, it introduces several hyperparameters to tune. We claim that it is not necessary to produce multimodal output. Contrary to the state-of-the-art approaches, we propose handling the conditional image-to-image problem in an entirely different approach from the unconditional case. 

\section{Methodology}
\subsection{Problem setup}
Our goal is to perform multimodal unsupervised image-to-image translation. Given an image \textit{x} from source domain \textbf{X}, we want to translate it to many images \textit{y\textsubscript{i}} in domain \textbf{Y}. To produce multimodal output, we accept a latent code  $ \textit{z} \sim \mathcal{N}(0\textsuperscript{k x 1},I\textsuperscript{k x k}) $
that is expected to describe the ways in which our output should differ. So our task is to find a function G such that: $ \text{G: }(\textit{x}, \textit{z\textsubscript{i}}) \rightarrow \textit{y\textsubscript{i},} $ where $ \textit{x} \in \textbf{X}, \textit{y\textsubscript{i}} \in \textbf{Y}  $. \par

\subsection{The latent scaling approach}
The core argument of our work presented here is that the latent code in conditional image generation should be interpreted differently than in the unconditional case. Traditionally, the latent code is concatenated with the input image directly or in feature space after compression through autoencoding. This follows the interpretation of the latent code as encoded data to be converted to a full-sized image. 
Instead, we propose interpreting the latent code entries as modulators of local changes in the input image. Specifically, the latent code is not considered as encoded data, but as a modifier of the convolutional operations of the network. A simple analogy is that previous methods generate diverse new images by appending the input image with different channels. On the other hand, our method generates diverse new images by using different brushes. We map the latent code to filters, such that the latent code modulates the strength of applying each filter. This is easiest to explain in the case of having a latent vector of length 3 operating only on the last 3 channels of the network. In that case, the latent vector will modulate only the color of the output image. However, in the hidden layers of the generator network, scaling the filters will correspond to modulating the resulting feature maps. For example, scaling an edge-detecting filter might result in stronger edges in the final image.
Given a k-dimensional latent code and an input image, we push the latent code through a fully-connected network to produce a scalar per filter. Then, the image is pushed through the convolutional network where each filter is scaled by the mapped latent code.
By not treating the latent code as data to be concatenated with the input image, we provide a simpler solution of the problem. Our approach allows a simple image-to-image GAN to achieve comparable quality with state-of-the-art while producing more diverse target domain images. Furthermore, without modifying standard GAN loss, we achieve disentanglement between source domain image content and target domain image style. This occurs as a result of our formulation where the latent code corresponds to local changes in the input image. 

\subsection{Simplifying hyperparameters for multimodality}
One major benefit of our method is the way it preserves the simple GAN loss. Since previous methods treat the latent code as compressed data, a simple concatenation of the latent code to the input image often leads to unimodal behavior where the latent code is ignored. This might be a result of augmenting a high-dimensional image with a low-dimensional vector. It is clear that the traditional usage of latent codes in unconditional GANs does not naturally extend to image-to-image GANs, as it requires adding multiple losses for reconstruction of images and latent codes. This is evident in the losses of BicycleGAN\cite{BicycleGAN} and MUNIT \cite{MUNIT}.

\begin{figure}[ht!]
\begin{center}
\includegraphics[width=1\linewidth]{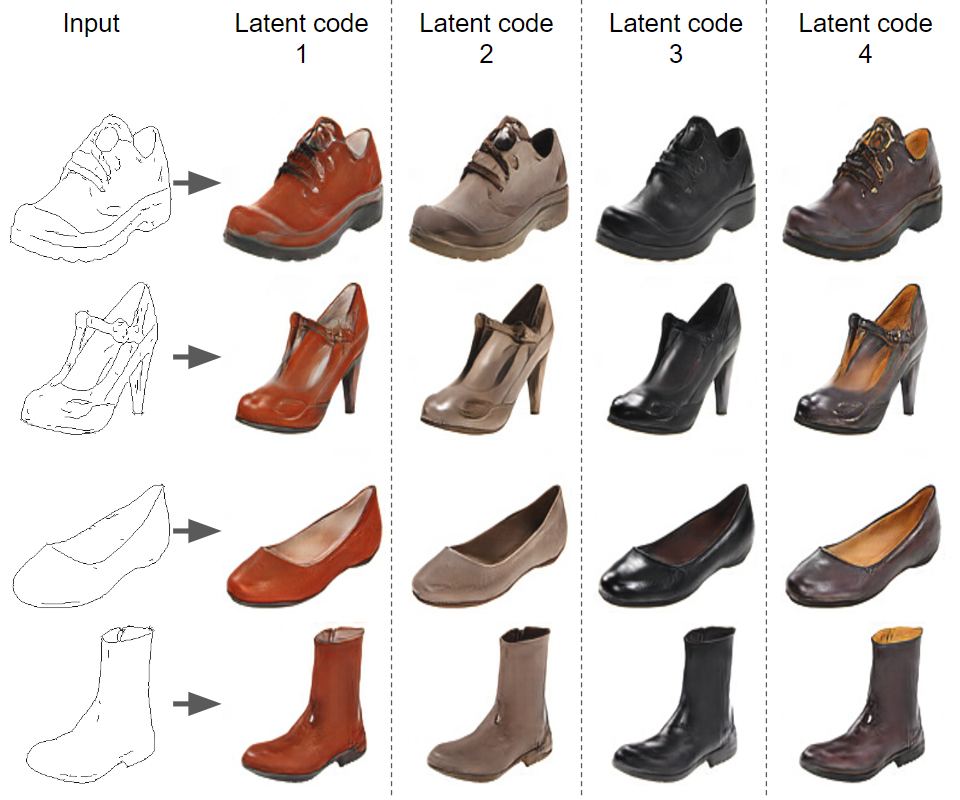}
\end{center}
   \caption{The results of applying the same latent codes to different input images. Disentanglement occurs as a result of our change of viewpoint and without explicit guidance.}
\label{fig:Disentanglement}
\end{figure}

\begin{figure}[H]
\begin{equation}
\begin{split}
&\mbox{Bicycle GAN loss:} \\
 \min_{\G,\E} \max_{\D} \quad  & \mathcal{L}_{\text{GAN}}^{\text{VAE}}(\G,\D,\E) + \lambda \mathcal{L}_1^{\text{VAE}}(\G,\E) \\
+& \mathcal{L}_{\text{GAN}}(\G,\D) \\ +& \lambda_{\text{latent}} \mathcal{L}_1^{\text{latent}}(\G,\E) + \lambda_{\text{KL}}\mathcal{L}_{\text{KL}}(\E)
\end{split}
\end{equation}
\end{figure}
Where $\lambda$ is a hyperparameter that controls the weight of the L1 VAE loss, $\lambda_{\text{latent}}$ controls the weight of the reconstruction loss of the latent code, and $\lambda_{\text{KL}}$ controls the weight of encouraging the encoded distribution to be similar to a random Gaussian.

\begin{figure}[H]
\begin{equation}
\begin{split}
&\mbox{MUNIT loss:} \\
\underset{E_{1}, E_{2}, G_{1}, G_{2}}\min&\ \underset{D_{1}, D_{2}}\max\ \mathcal{L}(E_{1}, E_{2}, G_{1}, G_{2}, D_{1}, D_{2}) = \mathcal{L}^{x_{1}}_{\text{GAN}} \\ +& \mathcal{L}^{x_{2}}_{\text{GAN}}\ + 
	\lambda_{x}(\mathcal{L}^{x_{1}}_{\text{recon}}+\mathcal{L}^{x_{2}}_{\text{recon}})\\+&\lambda_{c}(\mathcal{L}^{c_{1}}_{\text{recon}}+\mathcal{L}^{c_{2}}_{\text{recon}})+\lambda_{s}(\mathcal{L}^{s_{1}}_{\text{recon}}+\mathcal{L}^{s_{2}}_{\text{recon}})
	\label{equ:loss}
\end{split}
\end{equation}
\end{figure}
Where $\lambda_{\text{x}}$ controls the weight of image reconstruction loss, $\lambda_{\text{c}}$ controls the weight of disentangled content code reconstruction loss, and $\lambda_{\text{s}}$ controls the weight of disentangled style code reconstruction loss for both source and target domains. \par
\begin{figure}[ht!]
\begin{center}
 \includegraphics[width=1\linewidth]{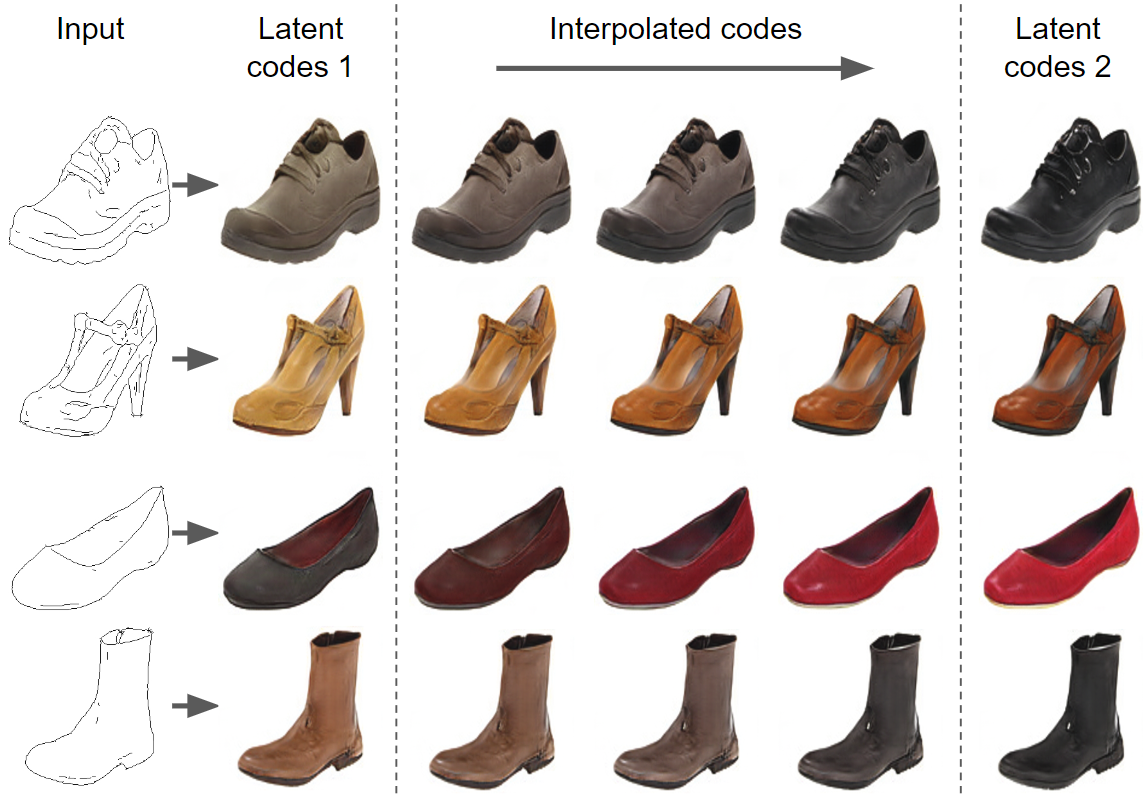}
\end{center}
   \caption{The results of interpolating between latent codes and generating the corresponding images. }
\label{fig:LatentWalk}
\end{figure}
By treating the latent code as a modifier of the network's filters, we can use the traditional GAN loss without any additional encoding or decoding, while preventing mode collapse for image-to-image translation. Our network learns to map an input Gaussian latent code to a scalar per filter, consequently learning to map different scalings of the network's filters to different output images. Throughout training, the latent code is never injected into the input image nor the learned feature maps. Instead, the latent code entries are multiplied by the feature maps. Consequently, we avoid the need for reconstruction losses and the need for losses based on encoded latent code diversity, and we use only the standard GAN loss. \par

\begin{figure}[H]
\begin{equation}
\begin{split}
&\mbox{Latent scaling loss:} \\
&\underset{G}\min\ \underset{D}\max = \mathcal{L}_{\text{GAN}}(\G,\D) 
	\label{equ:loss}
\end{split}
\end{equation}
\end{figure}

We choose to use the least-squares GAN loss (LSGAN) as described in \cite{LSGAN}:
\begin{figure}[H]
\begin{equation}
\label{eq:lsgan}
\begin{split}
\min_D \mathcal{L}_{\text{GAN}}(D) = &\mathbb{E}_{\bm{y} }\bigl[(D(\bm{y})-1)^2\bigr]\\&+ \mathbb{E}_{\bm{x}, \bm{z}}\bigl[D(G(x, \bm{z}))^2\bigr] \\
\min_G  \mathcal{L}_{\text{GAN}}(G) = &\mathbb{E}_{\bm{x}, \bm{z}}\bigl[(D(G(x, \bm{z}))-1)^2\bigr]
\end{split}
\end{equation}
\end{figure}
We apply label smoothing such that the desired value for discriminator real samples and for generator samples is 0.9 instead of 1.

\subsection{Mapping a low-dimensional latent vector to scalars}
There is a design choice in how to map the latent vector to the actual scalars that affect feature maps. One option is to sample a Gaussian with the same dimensionality as the network's filters. This leads to highly undesirable effects. As noted in \cite{BicycleGAN}, having a latent vector with high dimensionality makes sampling more difficult and leads to the network modeling less meaningful sources of variation. Since the total number of filters in state-of-the-art image-to-image networks is high (over 2000 filters), we use a fully-connected network to map a low-dimensional latent vector to the actual number of filters. \par
The mapping process leads to desirable effects. First, it allows us to maintain the same latent code sampling procedure as previous methods in the field. Since our method multiplies the filters by the latent code, we find that it helps training to learn how to scale the input latent code. Second, it allows the network to learn to make filters work in tandem. Finally, control over the final mapped scalars allows for controlling the tradeoff between quality and variability in output images.

\subsection{Disentanglement through mapped scalars}
After training, we examine the learned mapping between input noise which is sampled from a Gaussian distribution. We find that the latent codes, without explicit guidance, are style codes that are independent of the input image. Meaning that providing the same input code with multiple source domain images will produce target domain images of the same style. This indicates that our network is learning target domain style independently of the input image, and not correlating specific inputs with specific styles. Thus, our method obtains disentanglement between the source domain image and the target domain style essentially for free. Disentanglement occurs as a result of the network mapping the latent code to operations on the input images. Just as painting a shoe with a certain color or with certain specularity should easily transfer to other shoes. We show the result of using the same latent codes with different images in Figure \ref{fig:Disentanglement}.\par
Additionally, we demonstrate how interpolating between two latent codes leads to semantically interpolated images, as shown in Figure \ref{fig:LatentWalk}. The interpolation is not only in color, but also in features such as specularity and the presence of shoe string holes.

\begin{figure}[t!]
\begin{center}
 \includegraphics[width=1\linewidth]{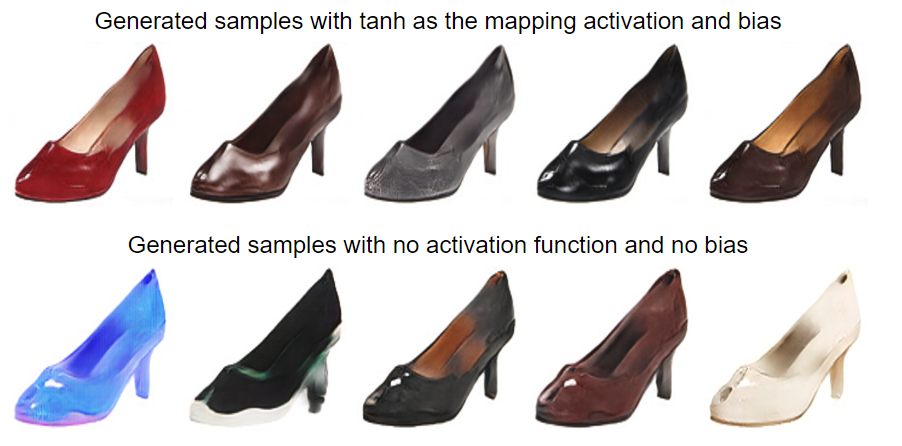}
\end{center}
   \caption{A demonstration of user control over the variability/quality tradeoff. An affine activation leads to more stylistic variety at the cost of lower quality images. A tanh activation leads to very realistic images but with lower variability. }
\label{fig:tradeoff}
\end{figure}

\subsection{Controlling the tradeoff between variability and quality}
The only tuning required by our method is tuning the mapping between input latent codes and the final scalars per filter. In our experiments, the mapping choice can be tuned easily. We find that applying the hyperbolic tangent function (tanh) as the activation function leads to conservative variation with the benefit of high quality images. Using leaky rectified linear unit (LRELU) or a linear fully-connected network leads to more variability at the cost of occasionally generating unrealistic images. This can be attributed to the magnitude of the mapped scalars being bounded when using tanh and unbounded when using LRELU or no activation.\par
We find that using bias in the fully-connected network leads to more conservative results. In this case, we believe that the average target domain image is contained in the bias, while the variation is achieved by the multiplied input code values. Eliminating bias leads to more variation in the produced images but again at the cost of quality.\par

Tuning the mapping between latent codes and scalars per filter allows user control over variety and quality of produced images. While there's dependence on the specific dataset, we find that using tanh in the final activation layer with bias leads to high quality low variety images. On the other hand, using an affine layer (linear activation without bias) leads to high variety low quality images. We show some results of modifying the scalar mapping in Figure \ref{fig:tradeoff}.

\subsection{Compatibility with existing deep learning libraries}
In many deep learning packages, it is easier to handle feature maps per batch than filters. This is because filters are per network, while feature maps are per input image. Therefore, we choose to multiply the latent vector entries by the feature maps instead of the filters to offer compatibility with most deep learning libraries. By the associative property of convolution, this is equivalent to the scalar multiplication with the filters:
\[ (c * f) \circledast I = c * (f \circledast I) \]
where c is a scalar, f is a k x k filter, I is a m x n image, * is the scalar multiplication operation, and \(\circledast\) is convolution. \par

\subsection{Implementation details}
We follow the network architecture described in \cite{CycleGAN} with a few differences as seen in Figure 1. First, our network contains only one generator that takes an image from the source domain and produces an image in the target domain, and one discriminator (instead of two generators and discriminators cycling between source and target domains). Second, we accept a latent code as an input to the network. Finally, we add a trainable fully-connected network to map the input latent code to scalars which are multiplied by each feature map of the network. The scalars are multiplied directly by the output of the convolution and before applying RELU or instance normalization. 
\begin{figure}[t!]
\begin{center}
 \includegraphics[width=1\linewidth]{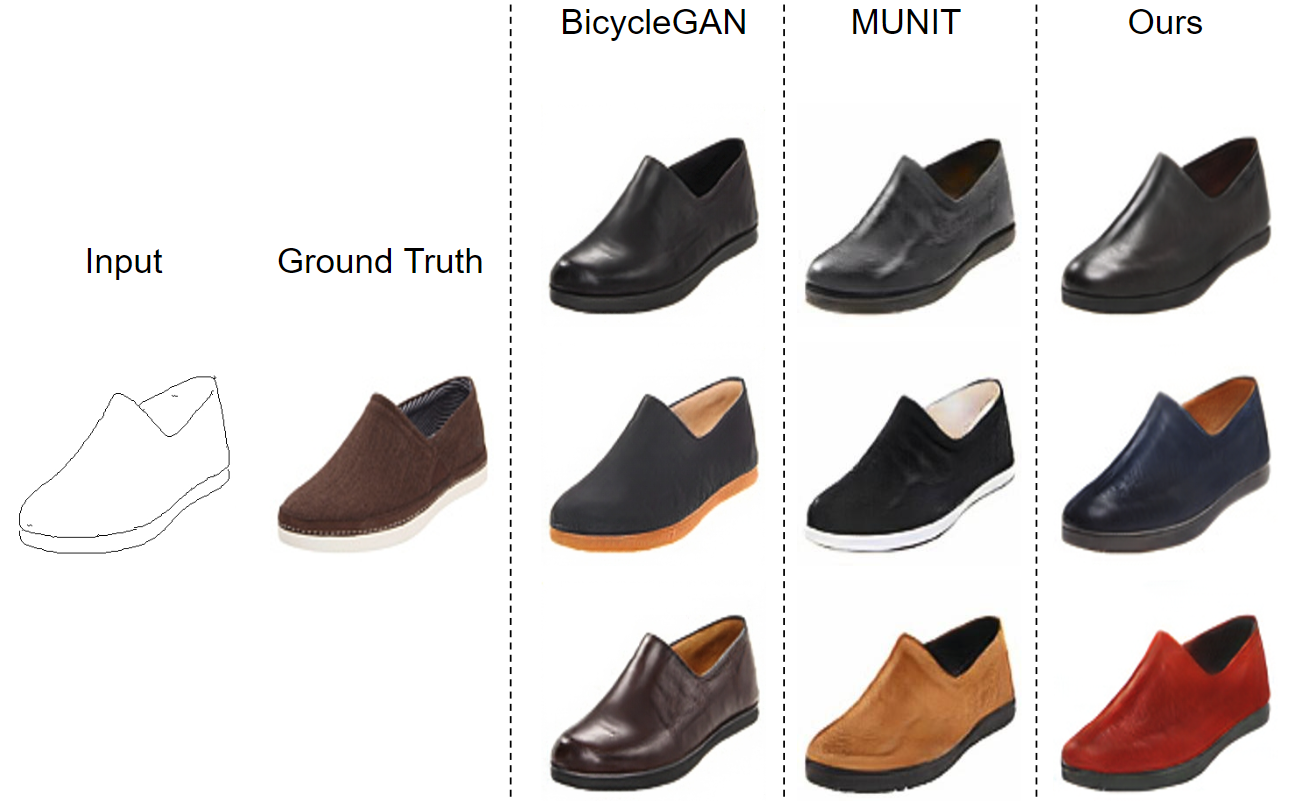}
\end{center}
   \caption{A comparison between shoes generated by our method, BicycleGAN, and MUNIT. AMT users preferred BicycleGAN to both, but preferred ours to MUNIT. }
\label{fig:shoeComparison}
\end{figure}

\subsection{Similarity to previous methods}
The idea of modifying the scale of a feature map was explored previously but with a different approach and for a different task. Adaptive Instance Normalization (AdaIN) \cite{adaIN} was proposed for style transfer, where the authors proposed transferring feature map statistics in order to transfer style from one image to another. A major difference between our method and AdaIN is that we learn feature map scales, where AdaIN simply computes scales from the input style image. Furthermore, we apply the scaling to feature maps prior to any normalization which is equivalent to filter scaling. StyleGAN is a more similar concurrent work where Karras \etal \cite{styleGAN} explore filter scaling and show great results for unconditional face image generation.




\section{Evaluation}
We perform qualitative and quantitative experiments using benchmark image-to-image translation datasets.  We show results on the following datasets: winter to summer \cite{SummerToWinter}, edges to shoes \cite{EdgesToShoes}, and labels to facades \cite{LabelsToFacades}. \par
For our qualitative experiments, we compare our results with BicycleGAN \cite{BicycleGAN}, and MUNIT \cite{MUNIT}. These are two state-of-the-art methods. BicycleGAN uses pair information, while MUNIT, as well as our method, do not use pair information. We use two metrics: quality, measured by Amazon Mechanical Turk (AMT) user preference, and diversity, measured by Learned Perceptual Image Patch Similarity (LPIPS) \cite{LPIPS}.

\begin{figure}[t!]
\begin{center}
 \includegraphics[width=1\linewidth]{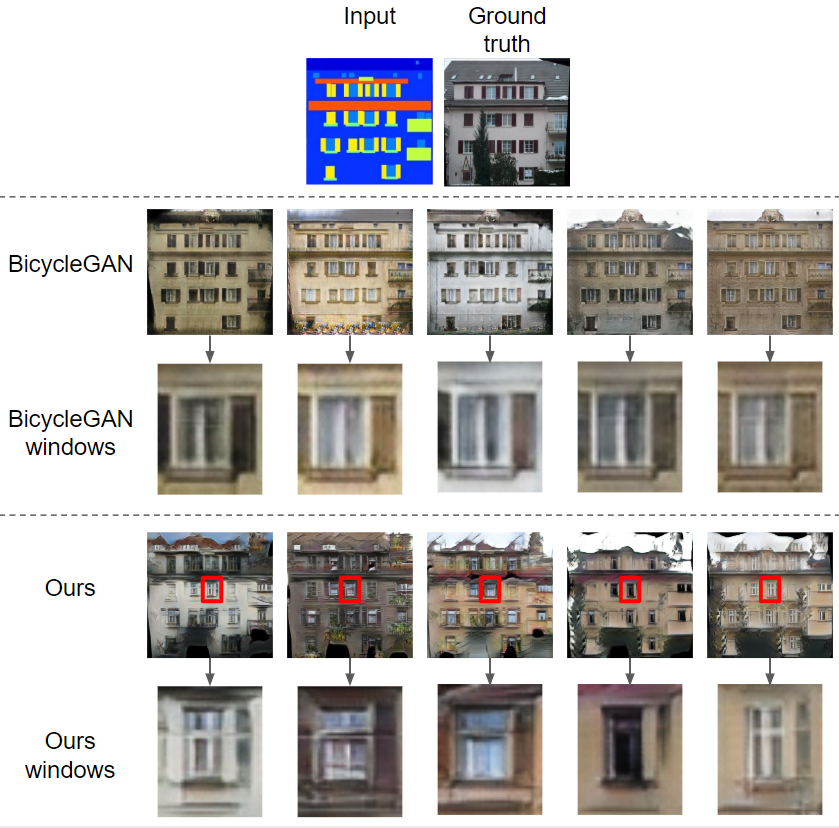}
\end{center}
   \caption{A comparison between the variety of generated windows in our method and that of BicycleGAN. Note how BicycleGAN is primarily changing color, while our method changes style as well.}
\label{fig:FacadesWindows}
\end{figure}

\subsection{Quantitative results}
We follow the same experimental settings as \cite{BicycleGAN} and \cite{MUNIT}. To measure quality, we present AMT users with two generated images: one by our algorithm, and the other by a different algorithm, and ask them to pick the image they prefer. We use the likelihood of images generated by an algorithm to be picked by users as the quality measure. Similarly to Bicycle GAN and MUNIT, we adopt the LPIPS \cite{LPIPS} diversity score as a quantitative measure of variability in the GAN output conditioned on the same image. We measure the LPIPS score using 1900 pairs, where each pair is two different generated images conditioned on the same input image. 
\begin{table}[H]

\centering 
\begin{tabular}{c c c} 
\hline\hline 
 & Quality & Diversity  \\ [0.5ex] 
\hline 
BicycleGAN & 57.2\% & 0.104  \\
[1ex] 
\hline 
\textbf{Ours} & --- & \textbf{0.109}  \\ 
[1ex] 
\hline 
MUNIT & 45.1\% & 0.109  \\ 
[1ex] 
\hline 

\end{tabular}
\caption{Comparison of quality and diversity between ours and state-of-the-art methods. Quality is measured by the percentage of images where another algorithm was preferred to ours. Diversity is measured by perceptual distance} 
\label{table:QuanResults} 
\end{table}

\begin{figure*}
\begin{center}
 \includegraphics[width=1\linewidth, height = 0.8\linewidth]{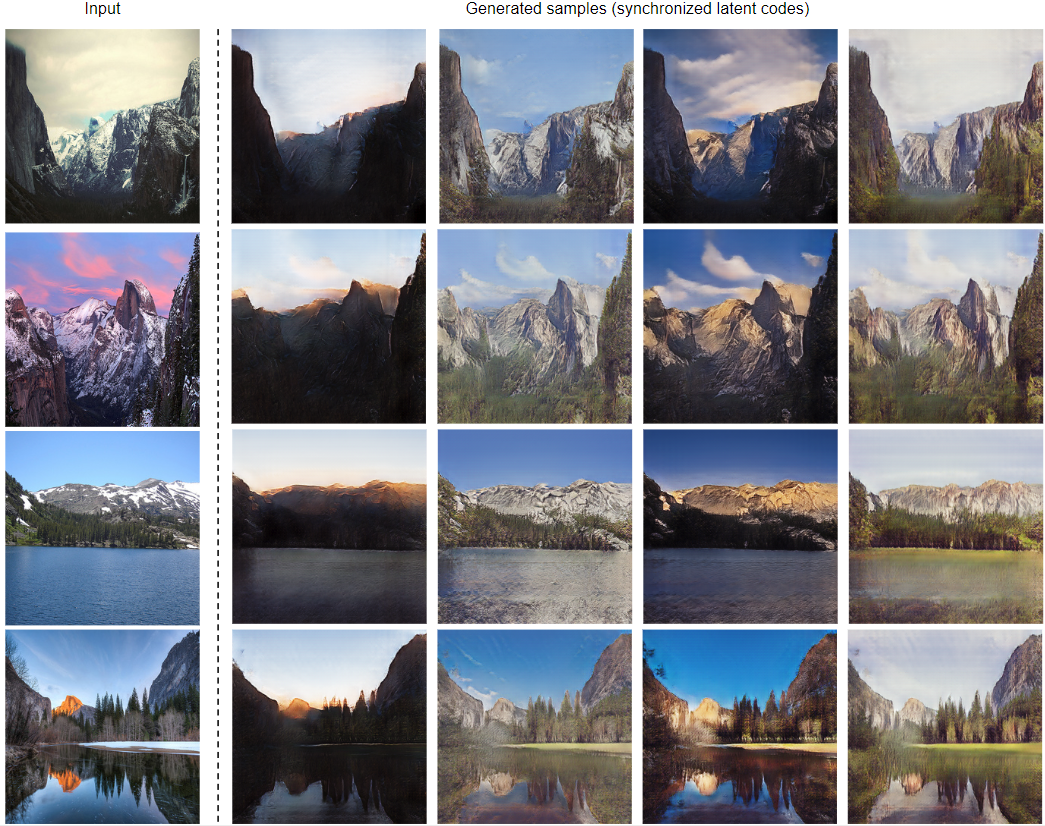}
\end{center}
   \caption{Our results on the winter to summer dataset. Learnt sources of semantic variation include time of day, presence of clouds, and amount of foliage.}
\label{fig:OursSummerWinter}
\end{figure*}

As shown in Table \ref{table:QuanResults}, our method results in the same variety (as measured by the pairwise LPIPS distance) as the best previously reported results but with higher AMT user rating of quality on the edges to shoes dataset. We confirm results by \cite{MUNIT} showing the superior quality of BicycleGAN which uses pair information. BicycleGAN quality is preferred to ours about 57\% of the time. This is comparable to the results obtained by MUNIT, showing that BicycleGAN is preferred to MUNIT 56\% of the time. In terms of diversity BicycleGAN scores lower than both our method and MUNIT.  However, we find that our method was preferred to MUNIT about 55\% of the time. The quality of images generated by our algorithm can be seen in \ref{fig:shoeComparison}. While our method uses only the standard GAN loss without additional hyperparameters, we obtain higher quality than MUNIT and more stylistic variability than BicycleGAN.


\subsection{Qualitative results}
We present images generated by our method to confirm the quantitative results in terms of variety and quality. Figure \ref{fig:OursSummerWinter} showcases our results on the winter to summer images. Since our method leads to style disentanglement, we produce synchronized results using the same code for each style. Our method learns several sources of semantic variability including time of day, condition of the sky, as well as amount of foliage. \par
As can be seen in Figure \ref{fig:FacadesWindows}, our method leads to more semantic variability which can be observed in the windows of generated facades. While BicycleGAN produces high-quality facade images, the variance in the window appearance is limited mainly to color. Our method on the other hand exhibits multiple realistic kinds of windows. This supports the higher diversity score of our method on the edges to shoes dataset. We believe that the use of pair information often restricts BicycleGAN to produce images that are too close to the ground truth. This explains why unsupervised methods such as our method and MUNIT produce images that exhibit more variety.\par

\subsection{Discussion}
Quantitative and qualitative results show that our method, while substantially simpler than both BicycleGAN and MUNIT, improves upon the best reported quality and diversity of upaired multimodal image-to-image GANs. We outperform MUNIT in terms of quality on the edges2shoes dataset while obtaining a higher diversity score than BicycleGAN. \par
Examining images generated by our method shows a high semantic variability. This includes generating different windows or architectural styles on the labels2facades dataset, and generating different times of day on the summer2winter dataset as seen in Figures \ref{fig:FacadesWindows} and \ref{fig:OursSummerWinter}. Generated images shown in Figure \ref{fig:shoeComparison} confirm our quantitative results. We find that, in general, BicycleGAN produces the highest quality images, while our method produces higher quality than MUNIT.\par
Additionally, after training using our method we show that latent codes are independent of the input images. As a result, latent codes are transferable from one output image to another, such that using the same latent code with different input images yields output images of the same style. This follows from our motivation where we map latent code entries to operations. Furthermore, interpolation between latent codes produces images that are interpolated in a semantic sense.\par
Our network, while simple and easy to tune, outperforms state-of-the-art methods. We believe that our success can be explained by referring to our change of viewpoint. Previous methods either extend unconditional GANs to accept images as additional input, or extent image-to-image GANs to accept a latent vector as additional input. As a result, autoencoding losses are needed to guarantee that input images and latent codes are not ignored. Therefore, a large portion of the training procedures is spent on tasks that are not necessary for multimodality, and only serves to preserve content after encoding or decoding. Our method, on the other hand, directly maps the latent code to the convolutional operations of the network. By design, the input image and the latent code will not be ignored. In addition, the latent code directly affects the convolutional operations. Thus, we avoid complicating the network architecture and we avoid solving auxiliary tasks. We believe that the simplicity and effectiveness of our method will lead to a wide adoption in the future in any image-to-image translation task. 

\section{Limitations and future work}
The main limitation of our method is finding a good mapping between randomly generated latent codes and scalars per filter. We believe that more work can be done in finding an optimal mapping. While almost any setting can lead to diverse and high-quality images, there are settings that are noticeably better than others.\par
Another limitation is related to disentanglement. Methods that use reconstruction losses to learn how to disentangle input images (such as MUNIT) have an advantage in style transfer. This is because the style from the input image can be applied to another input images. In our method the disentanglement occurs in the target domain. In other words, our method cannot extract style from an input image and apply it to another input image. The disentanglement in our method, however, can take a noise code that generated a certain output image and apply it to any input image to transfer the style. Thus, it is easy to transfer the style of a generated image to any input image, but to transfer the style of an input image to another input image the user must generate many images and use the latent code of the generated image most similar to the input image.

\section{Conclusion}
We present a method for multimodal unsupervised image-to-image translation. Our method is based on the idea that latent codes should be interpreted as modifiers of operations, and not as encoded data, in the case of conditional image generation. This formulation produces disentangled codes without autoencoding loss and without adding to the standard GAN loss. Our results show improvement on the state-of-the-art both qualitatively and quantitatively in terms of quality and diversity while using a drastically simpler network architecture. The simplicity of the architecture means easier implementation for users. In addition to simplicity, our method is general and can be applied to existing image-to-image translation methods such as CycleGAN \cite{CycleGAN}.
\par \textbf{Acknowledgement} The project was funded in part by the KAUST Office of Sponsored Research (OSR) under Award No. URF/1/3426-01-01.

{\small
\bibliographystyle{ieee}
\bibliography{egbib}
}

\end{document}